\documentclass[conference]{IEEEtran}
\usepackage{color}
\usepackage{textcomp}
\usepackage{amsmath}
\usepackage{graphicx}
\usepackage{wasysym}

\makeatletter
\ifCLASSINFOpdf
\else
\fi
\usepackage{amssymb}
\usepackage{url}

\makeatother

\begin{document}

\title{Auto-Encoder Guided GAN for Chinese Calligraphy Synthesis}


\author{\IEEEauthorblockN{Pengyuan Lyu\textsuperscript{1}, Xiang Bai\textsuperscript{1}, Cong Yao\textsuperscript{2}, Zhen Zhu\textsuperscript{1}, Tengteng Huang\textsuperscript{1}, Wenyu Liu\textsuperscript{1}} \IEEEauthorblockA{\textsuperscript{1}Huazhong University of Science and Technology, Wuhan, Hubei, China\\
\textsuperscript{2}Megvii Technology Inc., Beijing, China\\
{\tt\small \{lvpyuan, yaocong2010\}@gmail.com; \{xbai, jessemel, tengtenghuang, liuwy\}@hust.edu.cn}}}
\maketitle

\IEEEpeerreviewmaketitle{}
\begin{abstract}
In this paper, we investigate the Chinese calligraphy synthesis problem:
synthesizing Chinese calligraphy images with specified style from
standard font(\textit{eg.} Hei font) images (Fig. 1(a)). Recent works
mostly follow the stroke extraction and assemble pipeline which is
complex in the process and limited by the effect of stroke extraction.
We treat the calligraphy synthesis problem as an image-to-image translation
problem and propose a deep neural network based model which can generate
calligraphy images from standard font images directly. Besides, we
also construct a large scale benchmark that contains various styles
for Chinese calligraphy synthesis. We evaluate our method as well
as some baseline methods on the proposed dataset, and the experimental
results demonstrate the effectiveness of our proposed model.
\end{abstract}

\section{Introduction}

Chinese calligraphy is a very unique visual art and an important manifestation
of Chinese ancient culture which is popular with many people in the
world. Writing a pleasing calligraphy work is so difficult that it
always takes the writer many years to learn from the famous calligraphers'
facsimiles. Is there a way to synthesize calligraphy with specified
style expediently? We will explore an effective and efficient approach
for calligraphy synthesis in this paper.

Automatic calligraphy synthesis is a very challenging problem due
to the following reasons: 1) Various Chinese calligraphy styles. A
Chinese character usually has thousands of calligraphy styles which
vary from the shapes of component and the styles of strokes; 2) Deformations
between the standard font image and calligraphy image. The standard
font image and calligraphy image for the same character are only similar
in relative layout of radicals of character but different in the layout
and style of strokes.

Recently, there are some attempts \cite{xu2005automatic,xu2009automatic}
to synthesize calligraphy automatically, which first extract strokes
from some known calligraphy characters and then some strokes are selected
and assembled into a new calligraphy character. The above mentioned
methods are largely dependent on the effect of strokes extraction.
However, the stroke extraction technology does not always work well
when the Chinese character is too complex or the character is written
in a cursive style (Fig. 1(b)) where the strokes are hard to separate
and have to be extracted artificially \cite{xu2007intelligent}.

Considering there are some shortcomings in stroke assemble based methods,
we treat the calligraphy generation as an image-to-image translation
problem and propose a new method which can generate calligraphy with
a specified style from a standard Chinese font (\textit{i.e.} Hei
Font) directly without extracting strokes of characters. Over the
past few years, many network architectures have been proposed and
applied to different image-to-image tasks. However, those networks
are all designed to handle the pixel-to-pixel problems, such as semantic
segmentation, and poor performance is achieved when there are deformations
between the input and target images (Fig. 1(c)).

To overcome these problems, we propose a deep neural network based
model which consists of two subnets. The first one is an encoder-decoder
network acting as image transfer, which encodes an input standard
font image to a feature representation and then decodes the feature
representation to a calligraphy image with specified style. The encoder-decoder
network with similar architecture has been used in \cite{hinton2006reducing}
and show that the feature representation is likely to compress the
image content. This network architecture is sufficient to reconstruct
an image. But considering that in our task the input images and output
images only have the same relative layout among radicals but are different
in the layout and style of strokes, it is hard for an encoder-decoder
network to yield vivid calligraphy images. So besides the transfer
who captures the layout of input standard font image, we also use
another encoder-decoder network acting as autoencoder which inputs
and reconstructs calligraphy images to guide the transfer to learn
the detailed stroke information from autoencoder's low level features.
Finally, we train the two subnets together with reconstruct loss and
adversarial loss to make the output look real.

\begin{figure}
\begin{centering}
\includegraphics[scale=0.4]{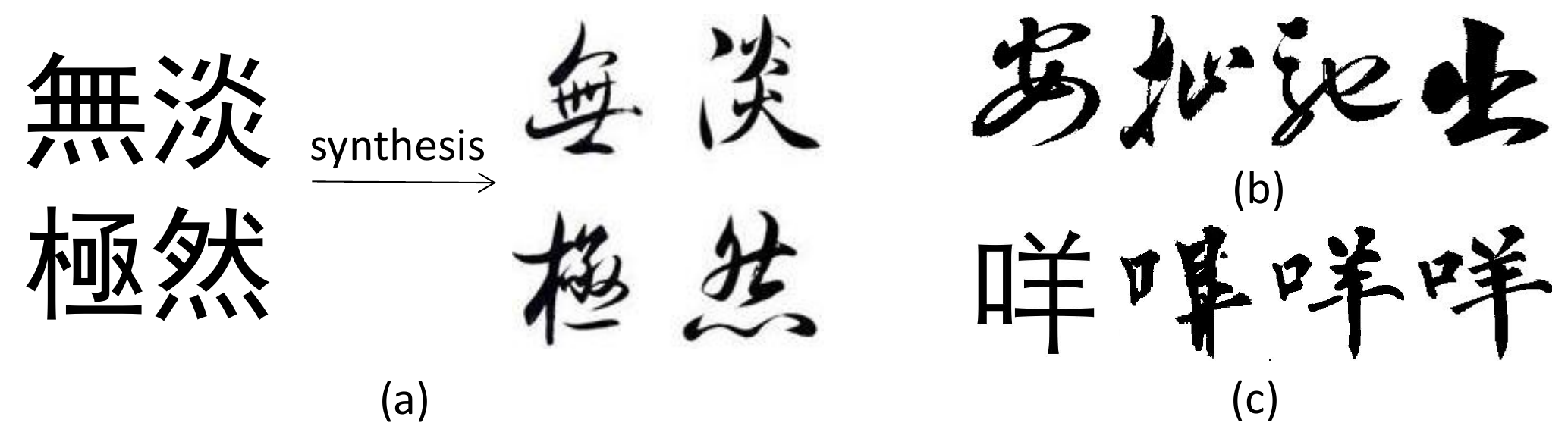}\caption{\label{fig_introdunction}(a) Synthesizing Chinese calligraphy images
from the standard font images. (b) Some calligraphy images in cursive
style whose strokes are hard to separate. (c) From left to right:
the standard font image, the output of UNet \cite{isola2016image},
the output of our method and the target image. The poor performance
is achieved by UNet while our method gets desired result. }
\par\end{centering}
\end{figure}

\begin{figure*}[!]
\centering{}\includegraphics[bb=35bp 0bp 1134bp 426bp,scale=0.45]{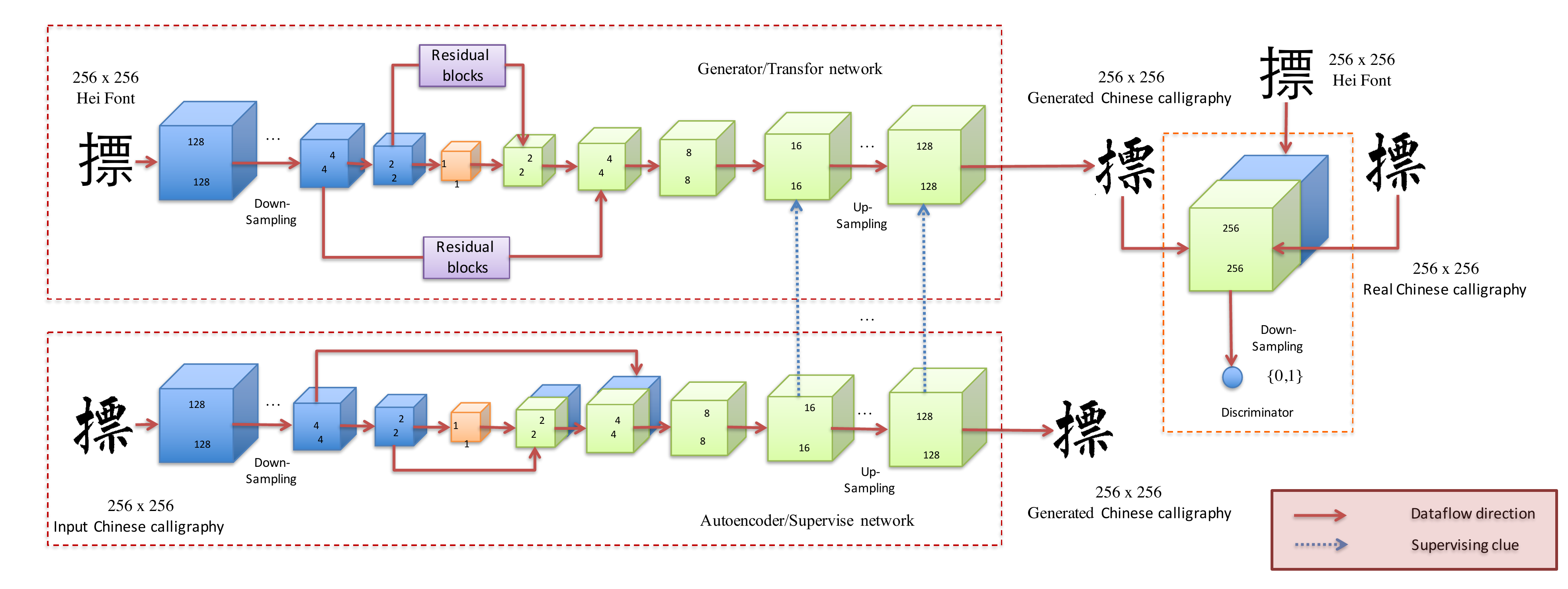}\caption{\label{fig_model}The architecture of the proposed method. The upper
part of image is our transfer network and the lower part of the architecture
is the supervise network which is an autoencoder. The whole network
is trained end-to-end and the supervise network is used to supervise
the low feature of transfer network's decoder in training phase.}
\end{figure*}

In summary, the contributions of this paper are two aspects: Firstly,
we propose a neural network based method which can end-to-end synthesize
calligraphy images with specified style from standard Chinese font
images. Compared to some baseline methods, our approach achieves the
best results with more realistic details. Secondly, we establish a
large-scale dataset for Chinese calligraphy synthesis collected from
the Internet. The dataset composes of 4 calligraphy styles and each
style contains about 7000 calligraphy images.

\section{Related Work\label{sec:Related-Work}}

\subsubsection{Chinese Calligraphy Synthesis}

In the past few years, many works on Chinese calligraphy synthesis
have been proposed. In \cite{xu2005automatic}, Xu et al. propose
a method based on shape analysis technique and hierarchical parameterization
to automatically generate novel artistically appealing Chinese calligraphy
artwork from existing calligraphic artwork samples for the same character.
Xu et al. \cite{xu2009automatic} propose a method to parameterize
stroke shapes and character topology, and successfully transfer font
Style Kai into a specific users\textquoteright{} handwriting style
by choosing the most appropriate character topology and stroke shapes
for a character. Different from the above mentioned methods which
follow the stroke extraction and stroke assembly pipeline, we input
a standard font image to our model and output a calligraphy image
directly.

\subsubsection{Image-to-Image Translation}

Image-to-image translation is an extensive concept which covers many
tasks such as edge/contour extraction \cite{xie15hed,shen2016object},
semantic segmentation \cite{long2015fully,noh2015deconv-segmantation},
artistic style transfer \cite{johnson2016perceptual,chen2016faststyletransfer},
image colorization \cite{luan2007natural,zhang2016colorful} et al.
in computer vision field. However, in those tasks, image-to-image
translation problems are often formulated as pixel-to-pixel translation
problem, where the input images and target images have the same underlying
structure and without any deformations. In this paper, we focus on
another scenario in image-to-image translation where there are deformations
between the input and target images. To be specific, in our calligraphy
synthesis task, the input standard font images and target calligraphy
images only have the similar relative layout among radicals of the
same characters but are different in the position and style of strokes.

\subsubsection{Generative Adversarial Networks}

Generative Adversarial Networks is proposed by \cite{goodfellow2014generative}
which has attracted great interest from the computer vision and machine
learning community and has a rapid development \cite{mirza2014cgan,radford2015dcgan,denton2015laplaciangan,chen2016infogan,arjovsky2017wgan}.
GAN is not only used in unsupervised learning such as generate an
image from random noise vector but also used with some image-to-image
translation tasks \cite{isola2016image,pathak2016context} to make
the output look real. Like \cite{isola2016image,pathak2016context},
we train our image transfer using an adversarial loss as well as the
reconstruction loss between the output images and target images to
generate desirable results. To learn the deformation between the input
images and target images,  we also reconstruct the low level features
of our transfer supervised by the low level feature from an autoencoder.

\section{Proposed Method\label{sec:Proposed-Method}}

In this section, we describe in detail the proposed method.  As shown
in Fig. \ref{fig_model}, our module consists of two encoder-decoder
networks which have similar network structure and can be trained together
in an end-to-end way. We refer to the two subnets as \textit{Supervise
Network} and \textit{Transfer Network} respectively, as \textit{Transfer
Network} is used to transfer a standard font image to a calligraphy
image with specified style, and \textit{Supervise Network} can provide
supervision information for \textit{Transfer Network} in training
stage. Details of\textcolor{blue}{{} }the two subnets are discussed
below.

\begin{figure}
\centering{}\includegraphics[bb=0bp 0bp 960bp 540bp,scale=0.25]{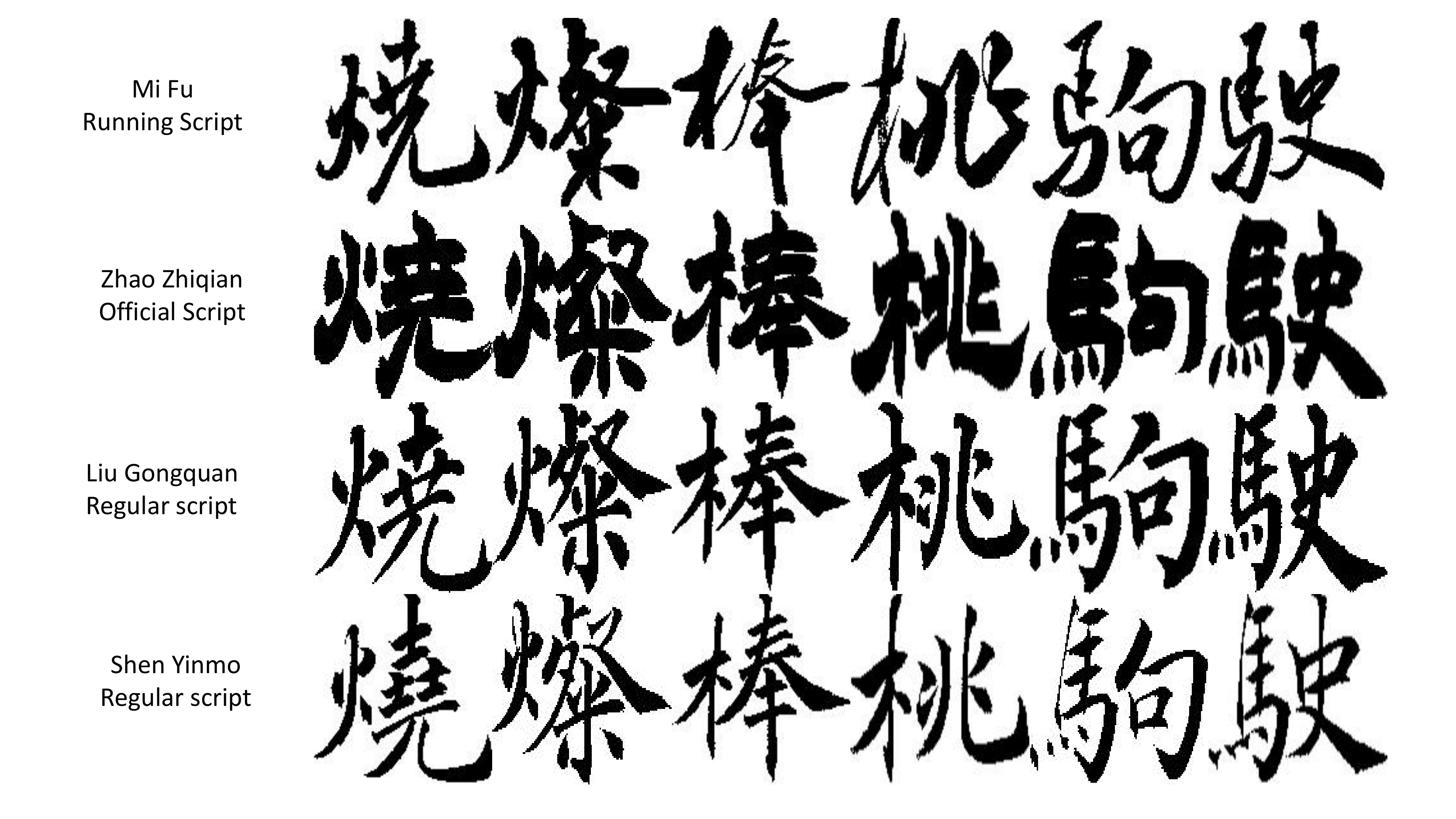}\caption{\label{fig_dataset}Some calligraphy images from the proposed benchmark.
This dataset contains 4 styles calligraphy images written by 4 famous
calligraphers in ancient China.}
\end{figure}

\subsection{Supervise Network}

The supervise network is an autoencoder network. The encoder consists
of a series of Convolution-BatchNorm-LeakyReLU \cite{radford2015dcgan}
blocks which takes a calligraphy image as input and produces a $C\times1\times1$
latent feature representation of that image, where $C$ is the dimension
of the latent feature. The decoder is stacked by a series of Deconvolution-BatchNorm-ReLU
\cite{radford2015dcgan} blocks, which takes the latent feature representation
from encoder and outputs an image which is similar to the input image. 

The architecture of the supervise network is a simple CNN based encoder-decoder
network but has skip connections between each layer $i$ and layer
$n-i$ as \cite{isola2016image}, where $n$ is the total number of
layers of supervise network. The skip connections are essential for
the supervise network to output images with photo-realistic details.
We have experimented and verified that the simple encoder-decoder
network can only output images with the rough layout but almost lost
all stroke information, but our supervise network can generate correct
strokes as input images. We argue that the feature maps of the bottleneck
layer in the simple encoder-decoder lost fine details of input images
but the spatial structure is kept, and that skip connections can provide
the decoder with detailed information. 

\subsection{Transfer Network}

The transfer network is also a CNN based encoder-decoder network which
inputs a standard font image and generates a calligraphy-like image.
The encoder and decoder are similar as the supervise network which
is composed by a series of Convolution-BatchNorm-LeakyReLU and Deconvolution-BatchNorm-ReLU
blocks respectively, but there is a little difference in skip connections. 

Chinese characters have diverse and complicated layouts and are hard
to transform to calligraphy image from standard font image even the
two images have the same layout. Instead of concatenating the feature
outputted by layer $i$ and layer $n-i$ directly, we use a residual
block \cite{he2016deep} to connect layer $i$ and layer $n-i$ and
sum the feature yielded by the residual block and layer $n-i$ to
enhance the capacity to learn the minute difference between the spatial
architecture of standard font and specified calligraphy images.

The standard font image and the corresponding calligraphy image always
have the same character component structure but vary greatly in the
layout and style of strokes. The high level features in encoder carry
layout information of the input standard font images, but it is not
enough to generate calligraphy images with clear strokes and specified
style when the model is only supervised by the target calligraphy
image. So we use the above supervise network to guide the transfer
network. Let $S=\{s_{1},s_{2},...,s_{k}\}$ and $T=\{t_{1},t_{2},...,t_{k}\}$
denote the low level feature representations yielded by supervise
network and transfer network's decoder respectively. We use $s_{j}$
to supervise $t_{j}$ in order to guide the decoder of transfer network
to learn the feature representations which carry the layout and style
of strokes, layer by layer.

Generative Adversarial Network (GAN) is recently proposed by \cite{goodfellow2014generative}
and has been widely used in image generation tasks \cite{larsen2015autoencoding,wu20163dlearning,zhang2016stackgan,isola2016image}.
Adversarial loss has the effect of learning the same distribution
of the ground truth distribution, which can make the output images
look real. \cite{isola2016image,pathak2016context} have shown that
an image transfer with an adversarial loss can output much sharper
results than the one only with L1 loss. We can adjust our transfer
network to a GAN framework easily with an additional discriminative
model. We treat transfer network as generator $G$ and use a deep
network as discriminative model $D$ following \cite{radford2015dcgan}.
In our work, the generator $G$ is optimized to output images which
have the same distribution as truth calligraphy images by generating
images that are difficult for the discriminator $D$ to differentiate
from real images. Meanwhile, $D$ is conditioned on the input standard
font images and optimized to distinguish real images and fake images
generated by $G$. 

\subsection{End-to-End Joint Training}

We train the two subnets jointly in an end-to-end way. Given a pair
of training sample $(x,y)$ which is composed of a standard font image
$x$ and a calligraphy image $y$ for the same character.

For the supervise network, we take calligraphy image $y$ as input
and the objective is to reconstruct $y$. We use L1 loss as our reconstruction
loss rather than L2 loss as L1 tends to yield sharper and cleaner
image. Let $A(.)$ be the supervise network, the objective of the
supervise network can be expressed as:
\begin{center}
$\mathcal{L}_{supervise}=\mathbb{E}_{y\AC p_{data}(y)}[||y-A(y)||_{1}]$
\par\end{center}

For the transfer network, we input standard font image $x$ and take
calligraphy font image $y$ as ground truth. We also reconstruct the
low level feature $S$ from the supervise network. We define the reconstruction
loss function as:
\begin{center}
$\mathcal{L}_{reconstruct-1}=\mathbb{E}_{t_{1}\AC p_{data}(t_{1})}[||t_{1}-s_{1}||_{1}]$
\par\end{center}

\begin{center}
...
\par\end{center}

\begin{center}
$\mathcal{L}_{reconstruct-(k)}=\mathbb{E}_{t_{(k)}\AC p_{data}(t_{(k)})}[||t_{(k)}-s_{(k)}||_{1}]$
\par\end{center}

\begin{flushleft}
$\mathcal{L}_{reconstruct}=\lambda_{1}\times\mathcal{L}_{reconstruct-1}+...+\lambda_{k}\times\mathcal{L}_{reconstruct-(k)}$
\par\end{flushleft}

Besides, we define our adversarial loss as:
\begin{flushleft}
$\mathcal{L}_{adversarial}=\mathbb{E}_{y\sim pdata(y)}[\log D(x,y)]+\mathbb{E}_{x\sim pdata(x)}[\log(1-D(x,G(x)))]$
\par\end{flushleft}

So, our final objective is:
\begin{flushleft}
$G^{*}=\arg\underset{G}{\min}\underset{D}{\max}\mathcal{L}_{adversarial}+\lambda_{s}\mathcal{L}_{supervise}+\lambda_{r}\mathcal{L}_{reconstruct}$
\par\end{flushleft}

\subsection{Implementation details}

In this paper, all images are scaled to 256 \texttimes{} 256 and converted
to binary images before being fed into the model. In addition, we
employ data augmentation to artificially enlarge the dataset for the
purpose of reducing overfitting. We flip the image horizontally with
probability of 0.5. 

The encoder of supervise network and transfer network both have 8
stacked Convolution-BatchNorm-LeakyReLU blocks, which yield 1 \texttimes{}
1 latent feature representations of the input calligraphy images and
standard font images respectively. The decoder of supervise network
and transfer network both have 7 stacked Deconvolution-BatchNorm-ReLU
blocks and followed by a Deconvolution layer which will generate 256
\texttimes{} 256 binary images. All Convolution and Deconvolution
layers in the above mentioned part have 4\texttimes 4 kernel size
and 2\texttimes 2 stride. The residual block in transfer net consists
of Convolution, Batch normalization and ReLU layers as \cite{he2016deep}
and only exists between the layers whose feature map size are $2\times2$
and $4\times4$ of encoder and decoder. The architecture of D is adapted
from \cite{radford2015dcgan}. 7 stacked Convolution-BatchNorm-ReLU
blocks are used and followed by a convolution layer and output the
probability of the input images like real. 

The method was implemented in Torch \cite{collobert2011torch7}. In
our experiment, we supervise the decoder of transfer net from the
layer with feature map size $16\times16$ to $128\times128$ and set
$\lambda_{1}...\lambda_{k}$ to 1, and set $\lambda_{s}$ and $\lambda_{r}$
to 100. We choose initial learning rate of 0.002 and train the proposed
model end-to-end with Adam \cite{kingma2014adam} optimization method.
This model was trained with batch size set to 16 until the output
tends to be stable in training phase. When testing, only the transfer
network is used to generate calligraphy images. We also use a median
filter to denoise the output image as a post-process method to make
the results cleaner.

\section{Experiments\label{sec:Experiments}}

In this section, we propose a new benchmark for Chinese calligraphy
generation and evaluate our algorithm on the proposed dataset. Besides,
we also compare our method with other neural network based image translation
methods to prove the effectiveness of our approach.

\subsection{Dataset}

As far as we know, there are no existing public datasets for Chinese
calligraphy images generation with specified style. Therefore we propose
a new dataset for calligraphy images automatic generation collected
from the Internet. This dataset contains 4 subsets which are written
by 4 famous calligraphers in ancient China, namely Mi Fu, Zhao Zhiqian,
Liu Gongquan and Shen Yinmo in different style. Some samples from
4 subsets are shown in Fig .\ref{fig_dataset}. What we can see is
that the styles of 4 subsets vary from one to another and cover a
few categories, such as running script, official script and regular
script. As shown, running script shows enormous shape transformation.
Official script exhibits wide and flat shapes. Its characters are
usually horizontally long and vertically short. Regular script is
more clear and neat which is mostly similar to printed fonts. Each
subset in our proposed benchmark contains about 7000 images and is
split into two set: training set and validation set. We randomly select
6000 images as training set and the rest images are treated as validation
set for each style and ensure that the training set and validation
set have no overlap in characters. 

For convenience, we call this dataset Chinese Calligraphy Synthesis-4(CCS-4).

\begin{figure}[t]
\centering{}\includegraphics[bb=0bp 240bp 960bp 1000bp,scale=0.55]{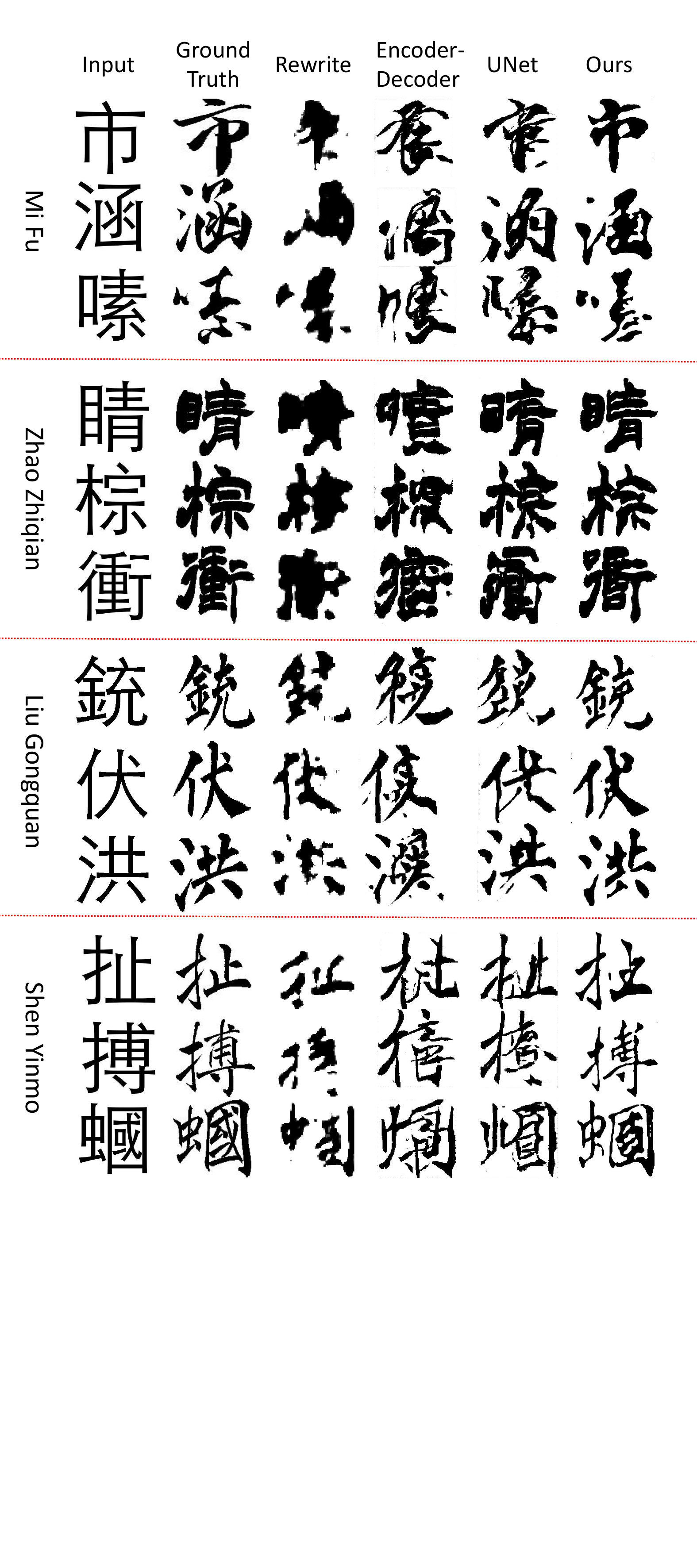}\caption{\label{fig_result} The results of the baseline methods as well as
our method on the 4 subset of the proposed benchmark.}
\end{figure}

\subsection{Baseline Methods}

\subsubsection{Rewrite }

Rewrite \cite{rewrite} is a neural style transfer for Chinese font
which is effective to transfer a typographic font to another stylized
typographic font. It is a simple top-down Convolution network with
big convolution kernel size and $1\times1$ stride. Each Convolution
layer in Rewrite is followed by Batch Normalization layer and a ReLu
layer. The architecture of Rewrite is stacked by some above mentioned
convolution blocks and end up with a $2\times2$ MaxPooling layer
then followed by a Dropout layer and a Sigmoid layer. The network
is minimized by L1 loss and total variation loss.

\subsubsection{Encoder-Decoder Network}

Encoder-Decoder network is an effective image-to-image translation
model and has been widely used and studied for many image translation
tasks, such as semantic segmentation \cite{noh2015deconv-segmantation},
edge extraction \cite{yang2016edge-extraction}, image colorization
\cite{isola2016image}, image restoration \cite{mao2016image-restoration}
and image style transfer \cite{chen2016faststyletransfer} \textit{etc}.
We use the architecture proposed by \cite{isola2016image} as a baseline
and train the model with L1 loss and adversarial loss.

\subsubsection{UNet}

UNet is proposed by \cite{ronneberger2015u} and is an extension of
the simple encoder-decoder method. In \cite{isola2016image}, skip
connections are used to connect encoder and decoder, based on the
fact that in an image translation task, the input and output differ
in surface appearance, but both are renderings of the same underlying
structure. Besides, the network is also optimized with L1 loss and
adversarial loss.

\subsection{Evaluation and Discussions}

We evaluate our proposed method as well as other baselines on the
CCS-4 dataset. We use the font style Hei as character\textquoteright s
standard font for each method, as Hei font has the least style structures,
even thickness and reduced curves. So using Hei font as character's
standard font can avoid the similarity between input font style and
output calligraphy style, which may increase the difficulty of calligraphy
generation but can evaluate the robustness and effectiveness of the
evaluated methods. 

\subsubsection{Qualitative Results}

We train a model for every above mentioned baseline methods as well
as our method on the four subsets individually. We show some samples
generated by the baseline methods and our proposed method in Fig.
\ref{fig_result}. Our method achieves the best results on all the
subsets. 

Specifically, Rewrite and Encoder-Decoder method achieve the worst
results. The images generated by Rewrite and Encoder-Decoder only
have the right spatial structure of Chinese character component but
the layout and style of strokes are far from satisfactory. As Fig.
\ref{fig_result} shows, In Mi Fu and Zhao Zhiqian subsets, almost
all strokes can not match the ground truth strokes.

The UNet method achieves a better result than the Encoder-decoder
result. Some results are close to ground truth both in global style
and local stroke style but have a small part of wrong strokes on the
Zhao Zhiqian, Liu Gong quan and Shen Yinmo subsets. However, the results
on Mi Fu subset is a little unsatisfactory. We argue that the layout
of strokes in Zhao Zhi qian, Liu Gongquan and Shen Yinmo are very
similar to the strokes of the standard font, which is much easier
to transform for the translation invariance of CNN. But there are
significant differences between the standard font and Mi Fu calligraphy
which may be too hard for UNet to learn.

The best results are obtained by our method. Even evaluated on Mi
Fu subset, our model can still generate images with the similar style
of the global image and the local stroke. In some scenes, such as
the input character has complex stroke structure, our method still
can handle well.

\subsubsection{Effect of The Supervise Network}

\begin{figure}
\includegraphics[bb=50bp 50bp 1000bp 350bp,scale=0.24]{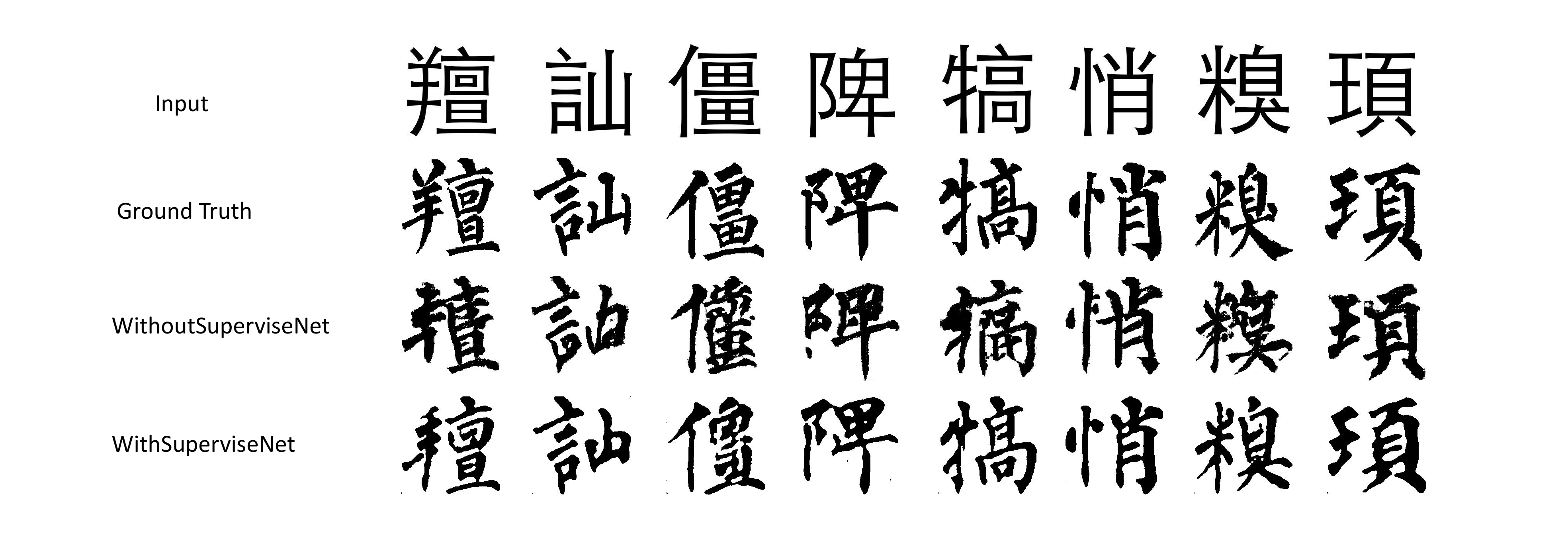}\caption{\label{fig_supervise}The results of our methods with/without supervise
network evaluated on Liu Gongquan subset.}
\end{figure}

In practice, low level features are hard to learn well in ordinary
autoencoder models, so we add Supervise Network in our model as a
reference to guide the transfer network to learn detail layout and
style of strokes. In Fig. \ref{fig_supervise}, we compare our model
with the one without Supervise network while other parts maintain
the same design. In the aspect of perceptual quality of the generated
font images, our model beats the one without Supervise network which
holds a general structure of the character while having some wrong
fine details.

\subsubsection{Effect of The Adversarial Loss}

\begin{figure}
\includegraphics[bb=50bp 50bp 1100bp 350bp,scale=0.24]{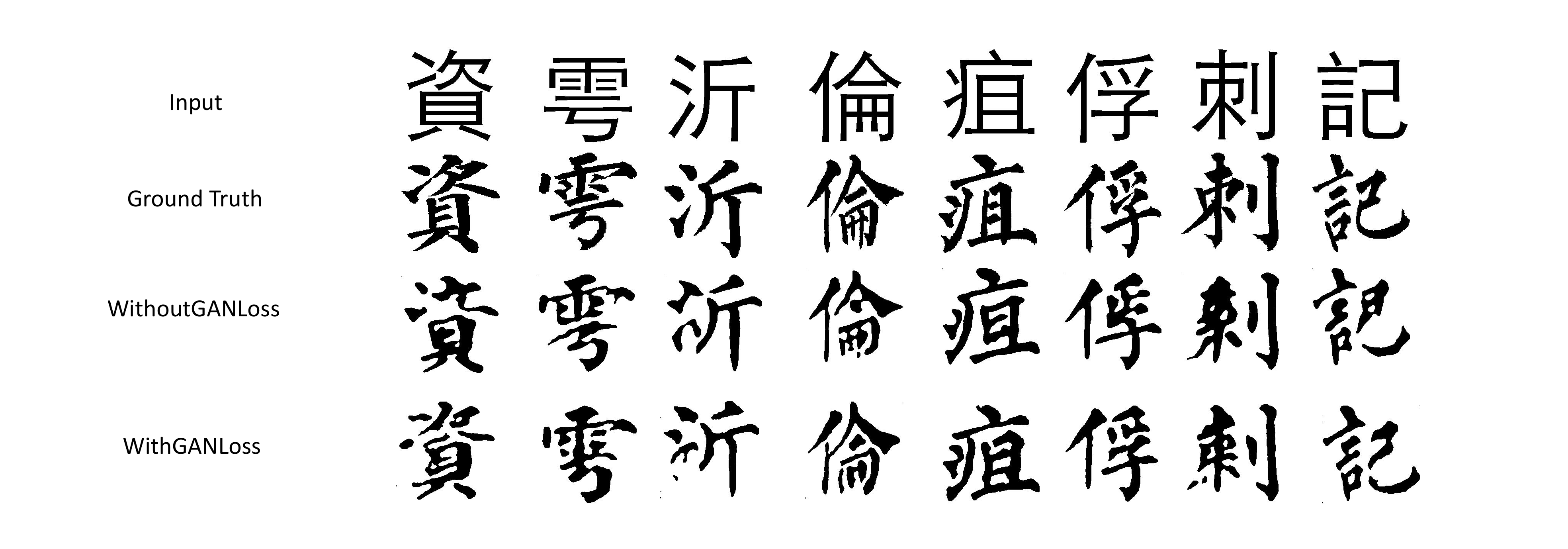}\caption{\label{fig_adversarial}The results of our methods with/without adversarial
loss evaluated on Liu Gongquan subset.}
\end{figure}

From the results shown in the Fig. \ref{fig_adversarial}, we can
see that the introduction of GAN Loss helps to improve the quality
of the generated image. It is obvious that there are more valid vivid
details of the characters are added and the generated image tends
to be more sharp with much less blur. Take the first column as example,
we can see that the generated image is similar with the ground image
in shape layout but loses some details. However, after adding GAN
Loss, the image generated is more sharp and detailed. We can draw
the conclusion that GAN Loss helps the generator mitigate the blur
and add the details which cannot be captured only with the L1 Loss.

\subsubsection{Analysis of The Standard Font}

\begin{figure}

\includegraphics[scale=0.4]{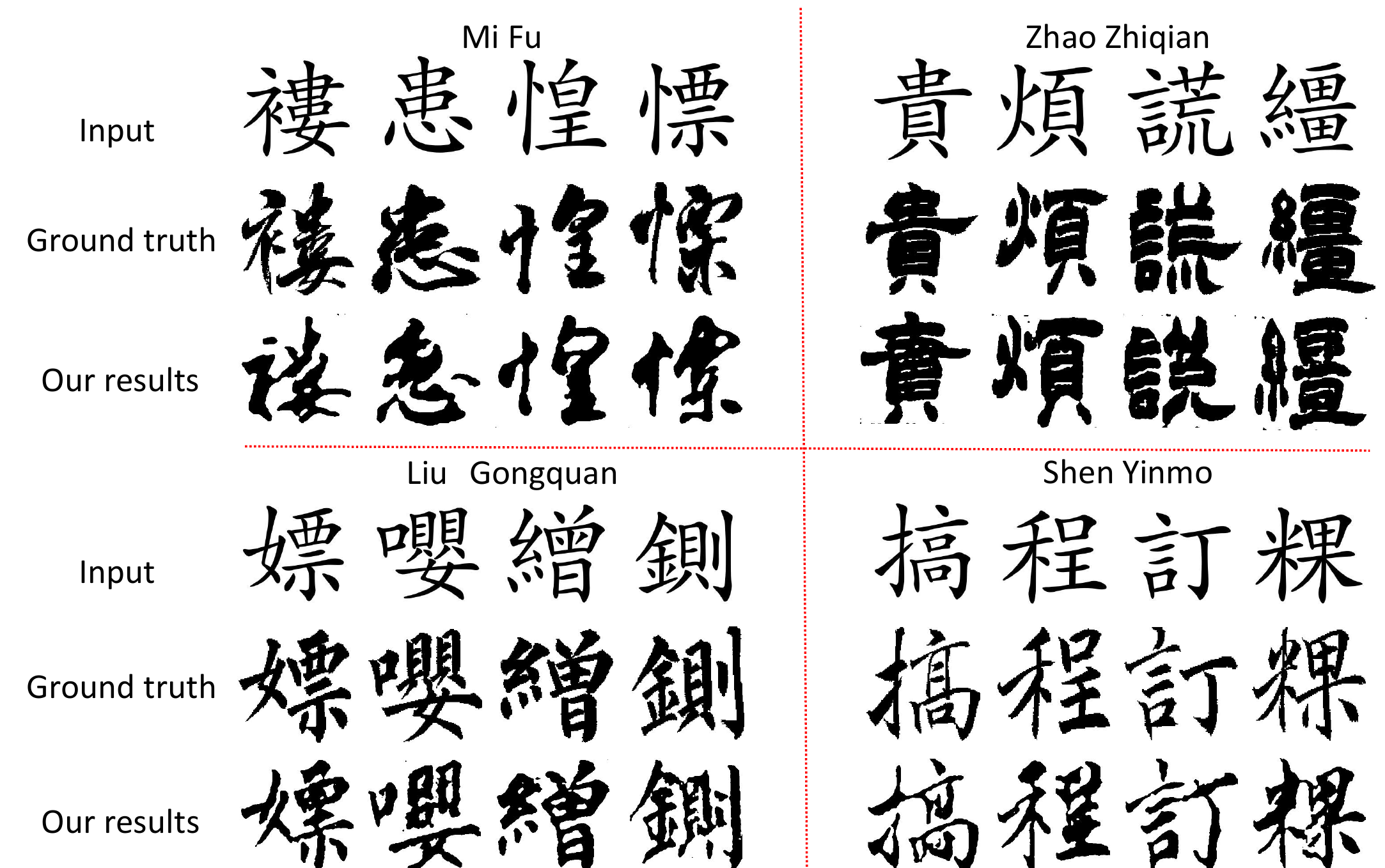}\caption{\label{fig:kaifont}The results of our method with Kai Font as standard
font. }

\end{figure}

Our approach achieves desirable results when we use Hei Font as the
standard font. Here, we use a different font as our standard font
to explore the affect of the standard font. As shown in Fig. \ref{fig:kaifont},
we use Kai font as the standard font, and our model can still output
photo-realistic calligraphy images which shows the robustness of our
method.

\section{Conclusion}

In this paper, we propose a model consisting of two subnets: \textit{transfer
network} and \textit{supervise network} to synthesize Chinese calligraphy.\textit{
}The transfer network can transfer a standard font image to a calligraphy
image with specified style and the supervise network can supervise
the transfer network to learn detailed stroke information. The two
subnets can be trained together. Compared with recent Chinese calligraphy
generation works, our approach can generate Chinese calligraphy images
from standard font images directly. Besides, we also establish a benchmark
for Chinese calligraphy generation and evaluate our method as well
as other baseline approaches on the proposed dataset. our method achieves
the best results. In the future, we will investigate methods that
can handle large deformation between the input target images and expand
our method to more general problems.

\bibliographystyle{plain}
\bibliography{icdar2017}

\end{document}